\documentclass[runningheads]{llncs}
\usepackage{graphicx}
\usepackage{amsmath,amssymb} 
\usepackage{color}\usepackage[width=122mm,left=12mm,paperwidth=146mm,height=193mm,top=12mm,paperheight=217mm]{geometry}

\newcommand{\hua}[1]{{\mathcal #1}}

\graphicspath{{./fig/}}

\usepackage{array}
\usepackage{subfigure}
\usepackage{epsfig}
\usepackage{graphicx}
\usepackage{amsmath}
\usepackage{amssymb}
\usepackage{bbm}
\usepackage{epstopdf}
\usepackage{caption}
\usepackage{enumitem}
\usepackage{calc}
\usepackage{multirow}
\usepackage{xspace}

\usepackage{booktabs}

\newcommand{\figref}[1]{Fig\onedot~\ref{#1}}

\newcommand{\tabref}[1]{Tab\onedot~\ref{#1}}

\newcommand{\ve}[1]{{\mathbf #1}} 

\makeatletter
\DeclareRobustCommand\onedot{\futurelet\@let@token\@onedot}
\def\@onedot{\ifx\@let@token.\else.\null\fi\xspace}
\def\eg{\emph{e.g}\onedot} 
\def\ie{\emph{i.e}\onedot} 
 
\def\etc{\emph{etc}\onedot} 
\def\wrt{w.r.t\onedot}

\usepackage{hyperref}
\usepackage{url}

\usepackage{array}
\usepackage{subfigure}
\usepackage{epsfig}
\usepackage{graphicx}
\usepackage{amsmath}
\usepackage{amssymb}
\usepackage{bbm}
\usepackage{epstopdf}
\usepackage{caption}
\usepackage{enumitem}
\usepackage{calc}
\usepackage{multirow}
\usepackage{xspace}

\usepackage{booktabs}

\begin{document}
\pagestyle{headings}
\mainmatter
\def\ECCV16SubNumber{0000}  

\title{Zoom Better to See Clearer: Human and Object Parsing with Hierarchical Auto-Zoom Net}
\titlerunning{ECCV-16 Submission}
\author{Fangting Xia, Peng Wang, Liang-Chieh Chen \& Alan L. Yuille}
\institute{University of California, Los Angeles\\
\texttt{\{sukixia,jerrykingpku,lcchen,yuille\}@ucla.edu}}
\maketitle
\begin{abstract}
Parsing articulated objects, \eg humans and animals, into semantic parts (\eg body, head and arms, \etc) from natural images is a challenging and fundamental problem for computer vision. A big difficulty is the large variability of scale and location for objects and their corresponding parts. Even limited mistakes in estimating scale and location will degrade the parsing output and cause errors in boundary details. To tackle these difficulties,  we propose a ``Hierarchical Auto-Zoom Net'' (HAZN) for object part parsing which adapts to the local scales of objects and parts. HAZN is a sequence of two ``Auto-Zoom Nets" (AZNs), each employing fully convolutional networks that perform two tasks: (1) predict the locations and scales of object instances (the first AZN) or their parts (the second AZN); (2) estimate the part scores for predicted object instance or part regions. Our model can adaptively ``zoom'' (resize) predicted image regions into their proper scales to refine the parsing. We conduct extensive experiments over the PASCAL part datasets on humans, horses, and cows. For humans, our approach significantly outperforms the state-of-the-arts by $5\%$ mIOU and is especially better at segmenting small instances and small parts. We obtain similar improvements for parsing cows and horses over alternative methods. In summary, our strategy of first zooming into objects and then zooming into parts is very effective. It also enables us to process different regions of the image at different scales adaptively so that, for example, we do not need to waste computational resources scaling the entire image.
\end{abstract}

\section{Introduction}
When people look at natural images, they often first locate regions that contain objects, and then perform the more detailed task of object parsing, \ie decomposing each object instance into its semantic parts. Object parsing, of humans and horses, is important for estimating their poses and understanding their semantic interactions with others and with the environment.

In computer vision, object parsing plays a key role for real understanding of objects in  images and helps for many visual tasks, e.g., segmentation~\cite{eslami2012generative,wang2015joint},  pose estimation~\cite{dong2014towards}, and fine-grained recognition~\cite{zhang2014part}. It also has many industrial applications such as robotics and image descriptions for the blind.

There has been a growing literature on the related task of object semantic segmentation due to the availability of
evaluation benchmarks such as PASCAL VOC~\cite{everingham2014pascal} and MS-COCO~\cite{lin2014microsoft}. There has been work on human parsing, \ie segmenting humans into their semantic parts, but this has mainly studied under constrained conditions which pre-suppose known scale, fairly accurate localization, clear appearances, and/or relatively simple poses~\cite{bo2011shape,zhu2011max,eslami2012generative,yamaguchi2012parsing,dong2014towards,LiuCVPR15}. There are few works done on parsing animals, like cows and horses, and these often had similar restriction, e.g., roughly known size and location~\cite{wang2014semantic,wang2015joint}.

In this paper we address the task of parsing objects, such as humans and horses, in ``the wild" where there are large variations in scale, location, occlusion, and pose. This motivates us to work with PASCAL images~\cite{everingham2014pascal} because these were chosen for studying multiple visual tasks, do not suffer from dataset design bias~\cite{li2014secrets}, and include large variations of objects, particularly of scale. Hence parsing humans in PASCAL is considerably more difficult than in datasets, such as Fashionista~\cite{yamaguchi2012parsing}, that were constructed solely to evaluate human parsing.

Recently, deep learning methods have led to big improvements on object parsing~\cite{hariharan2014hypercolumns,wang2015joint}. These improvements are due to fully convolutional nets (FCNs)~\cite{long2014fully} and the availability of object part annotations on large-scale datasets, e.g. PASCAL~\cite{chen2014detect}. Although these methods worked well, they can make mistakes on small or large scale objects and, in particular, they have no mechanism to adapt to the size of the object.

\begin{figure}[t!]
\begin{center}
   \includegraphics[width=1.0\linewidth]{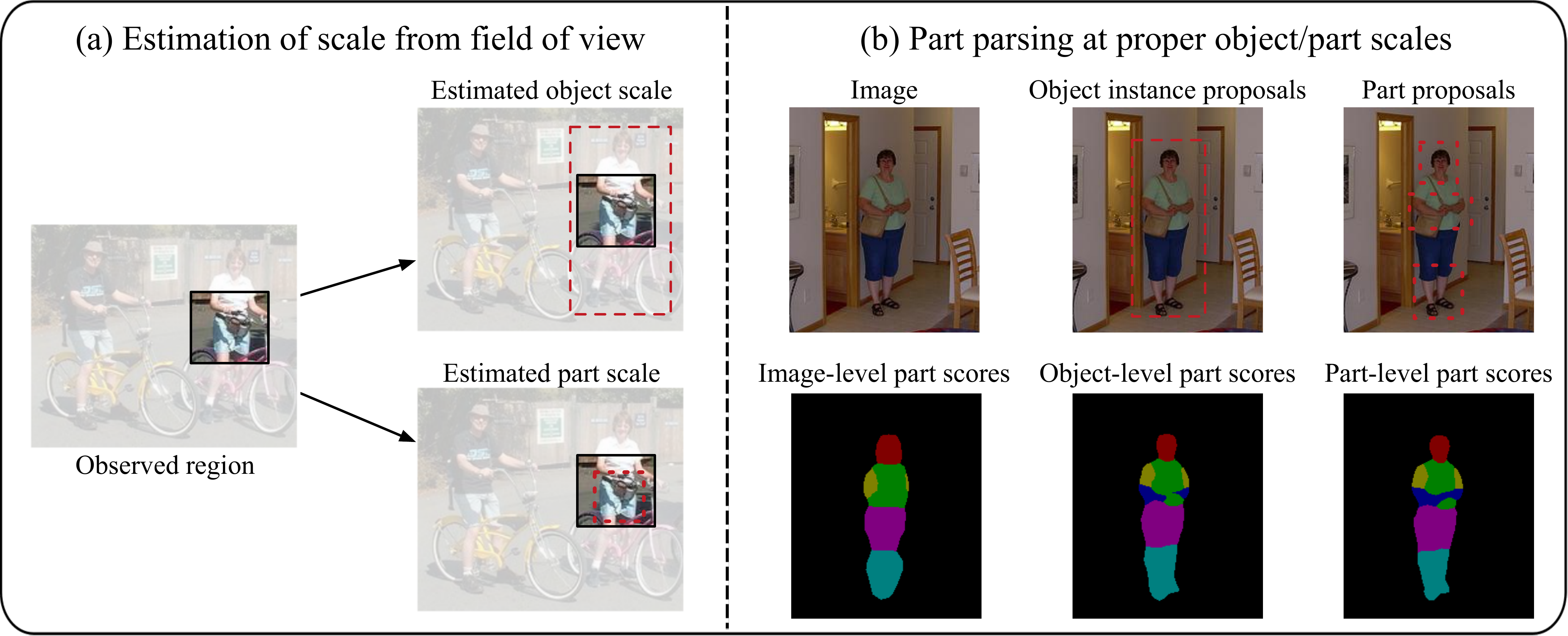}
\end{center}
\vspace{-1\baselineskip}
\caption{Intuition of our Hierarchical Auto-Zoom model (HAZN). (a) The scale and location of an object and its parts (the red dashed boxes) can be estimated from the observed field of view (the black solid box) of a neural network. (b) Part parsing can be more accurate by using proper object and part scales. 
At the top row, we show our estimated object and part scales. 
In the bottom row, our part parsing results gradually become better by increasingly utilizing the estimated object and part scales.}
\vspace{-1.6\baselineskip}
\label{fig:intuition}
\end{figure}

\begin{figure}[t]
\begin{center}
   \includegraphics[width=1.0\linewidth]{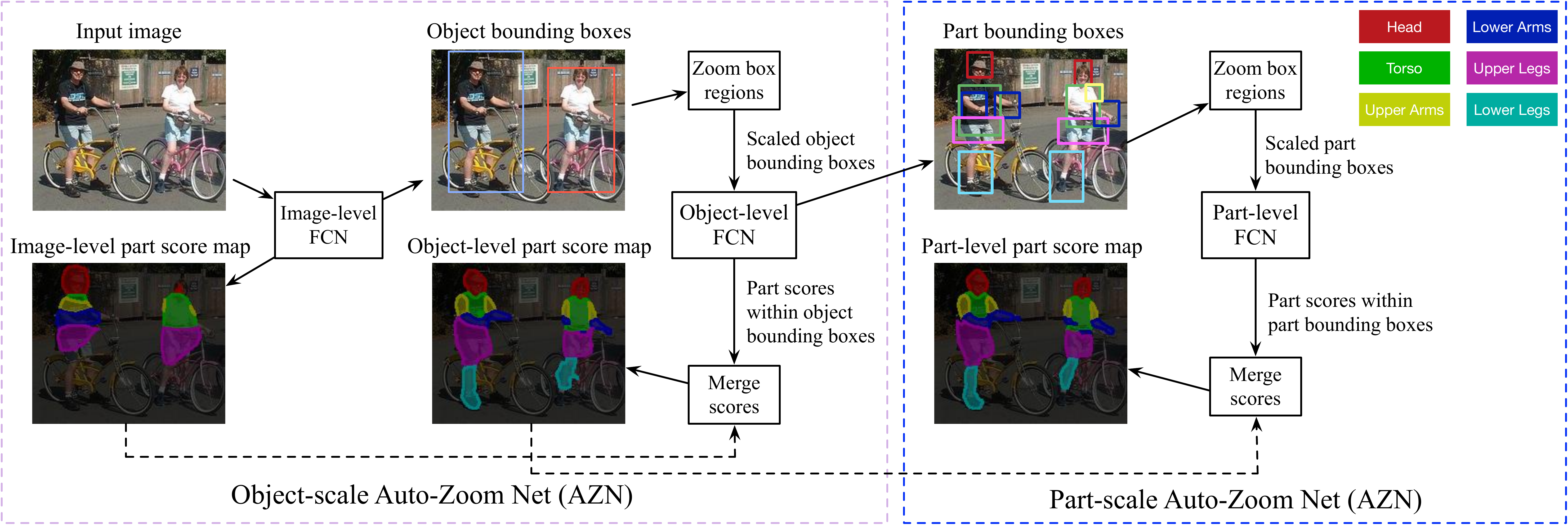}
\end{center}
\vspace{-1\baselineskip}
\caption{Testing framework of Hierarchical Auto-Zoom Net (HAZN). We address object part parsing in a wild scene, adapting to the size of objects (object-scale AZN) and parts (part-scale AZN). The part scores are predicted and refined by three FCNs, over three levels of granularity, \ie image-level, object-level, and part-level. At each level, the FCN outputs the part score map for the current level, and estimates the locations and scales for next level. The details of parts are gradually discovered and improved along the proposed auto-zoom process (\ie location/scale estimation, region zooming, and part score re-estimation).}
\vspace{-1.6\baselineskip}
\label{fig:framework}
\end{figure}

In this paper, we present a hierarchical method for object parsing which performs scale estimation and object parsing jointly and is able to adapt its scale to objects and parts. It is partly motivated by the proposal-free end-to-end detection strategies~\cite{huang2015densebox,ren2015faster,DBLP:journals/corr/RedmonDGF15,DBLP:journals/corr/LiangWSYLY15}.
To get some intuition for our approach observe, in Fig.~\ref{fig:intuition}(a), that the scale and location of a target object, and of its corresponding parts, can be estimated accurately from the field-of-view (FOV) window by applying a deep net.
We call our approach ``Hierarchical Auto-Zoom Net" (HAZN) which parses the objects at three levels of granularity, namely image-level, object-level, and part-level, gradually giving clearer and better parsing results, see Fig.~\ref{fig:intuition}(b). The HAZN sequentially combines two ``Auto-Zoom Nets" (AZNs), each of which predicts the locations and scales for objects (the first AZN) or parts (the second AZN), properly zooms (resizes) the predicted image regions, and refines the object parsing result for those image regions (see Fig~\ref{fig:framework}). The HAZN uses three fully convolutional neural networks (FCNs)~\cite{long2014fully} that share the same structure. The first FCN acts directly on the image to estimate a finite set of possible locations and sizes of objects (e.g., bounding boxes) with confidence scores, together with a part score map of the image. The part score map (in the bottom left of Fig.~\ref{fig:intuition}(b)) is similar to that proposed by previous deep-learned methods. The object bounding boxes are scaled to a fixed size by zooming in or zooming out (as applicable) and the image and part score maps within the boxes are also scaled (e.g.,by bilinear interpolation for zooming in or downsampling for zooming out). Then the second FCN is applied to the scaled object bounding boxes to make proposals (bounding boxes) for the positions and sizes of the parts, with confidence values, and to re-estimate the part scores within the object bounding boxes. This yields improved part scores, see the bottom-middle score map of Fig.~\ref{fig:intuition}(b). We then apply the third FCN to the scaled part bounding boxes to produce new estimates of the part scores and to combine all of them (for different object and part bounding boxes) to output final part scores, see the bottom right of Fig.~\ref{fig:intuition}(b), which are our parse of the object. This strategy is modified slightly so that, for example, we scale humans differently depending on whether we have detected a complete human or only the upper part of a human, which can be determined automatically from the part score map.

We now briefly comment on the advantages of our approach for dealing with scale and how it differs from more traditional methods. Previous methods mainly select a fixed set of scales in advance and then perform fusion on the outputs of a deep net at different layers. Computational requirements mean that the number of scales must be small and it is impractical (due to memory limitations) to use very fine scales. Our approach is considerably more flexible because we adaptively estimate scales at different regions in the image which allows us to search over a large range of scales. In particular, we can use very fine scales because we will probably only need to do this within small image regions. For example, our largest zooming ratio is 2.5 (at part level) on PASCAL while that number is 1.5 if we have to zoom the whole image. This is a big advantage when trying to detect small parts, such as the tail of a cow, as is shown by the experiments. In short, the adaptiveness of our approach and the way it combines scale estimation with parsing give novel computational advantages.

We illustrate our approach by extensive experiments for parsing humans on the PASCAL-Person-Part dataset~\cite{chen2014detect} and for parsing animals on a horse-cow dataset~\cite{wang2014semantic}. These datasets are challenging because they have large variations in scale, pose, and location of the objects. Our zooming approach outperforms previous state of the art methods by a large margin. We are particulary good at detecting small object parts.

\section{Background}
The study of human part parsing has been largely restricted to constrained environments, where a human instance in an image is well localized and has a relatively simple pose like standing or walking~\cite{bo2011shape,zhu2011max,eslami2012generative,yamaguchi2012parsing,dong2014towards,LiuCVPR15,DBLP:journals/corr/XiaZWY15}.
These works, though useful for parsing a well cropped human instance from simple commercial product images, are limited when applied to parsing human instances in the wild, since humans in real-world images are often in various poses, scales, and may be occluded or highly deformed. The high flexibility of poses, scales and occlusion patterns is difficult to handle by shape-based and appearance-based models with hand-crafted features or bottom-up segments.

Over the past few years, with the powerful deep convolutional neural networks (DCNNs)~\cite{lecun1998gradient} and big data, researchers have made significant performance improvement for semantic object segmentation in the wild~\cite{chen2014semantic,dai2015boxsup,liu2015semantic,noh2015learning,papandreou2015weakly,wang2015towards,tsogkas2015semantic}, showing that DCNNs can also be applied to segment object parts in the wild. These deep segmentation models work on the whole image, regarding each semantic part as a class label. But this strategy suffers from the large scale variation of objects and parts, and many details can be easily missed.
\cite{hariharan2014hypercolumns} proposed to sequentially perform object detection, object segmentation and part segmentation, in which the object is first localized by a RCNN~\cite{girshick2014rich}, then the object (in the form of a bounding box) is segmented by a fully convolutional network (FCN)~\cite{long2014fully} to produce an object mask, and finally part segmentation is performed by partitioning the mask.
The process has two potential drawbacks: (1) it is complex to train all components of the model; (2) the error from object masks, \eg local confusion and inaccurate edges, propagates to the part segments. Our model follows this general coarse-to-fine strategy, but is more unified (with all three FCNs employing the same structure) and more importantly, we do not make premature decisions.
In order to better discover object details and use object-level context, \cite{wang2015joint} employed a two-stream FCN to jointly infer object and part segmentations for animals, where the part stream was performed to discover part-level details and the object stream was performed to find object-level context. Although this work discovers object-level context to help part parsing, it only uses a single-scale network for both object and part score prediction, where small-scale objects might be missed at the beginning and the scale variation of parts still remains unsolved.

Many studies in computer vision has addressed the scale problem to improve recognition or segmentation. These include exploiting multiple cues~\cite{DBLP:journals/ijcv/HoiemEH08}, hierarchical region grouping~\cite{arbelaez2011contour,florack1996gaussian}, and applying general or salient object proposals combined with iterative localization~\cite{DBLP:journals/pami/AlexeDF12,Yukun_CVPR15,DBLP:conf/cvpr/WangWZFZL12}.
However, most of these works either adopted low-level features or only considered constrained scene layouts, making it hard  to handle wild scene variations and difficult to unify with DCNNs. Some recent works try to handle the scale issue within a DCNN structure. They commonly use multi-scale features from intermediate layers, and perform late fusion on them~\cite{long2014fully,hariharan2014hypercolumns,chen2014semantic} in order to achieve scale invariance. Most recently, ~\cite{chen2015attention} proposed a scale attention model, which learns pixel-wise weights for merging the outputs from three fixed scales. These approaches, though developed on powerful DCNNs, are all limited by the number of scales they can select and the possibility that the scales they select may not cover a proper one. Our model avoids the scale selection error by directly regressing the bounding boxes for objects/parts and zooming the regions into proper scales. In addition, this mechanism allows us to explore a broader range of scales, contributing a lot to the discovery of missing object instances and the accuracy of part boundaries.



\section{The Model}
As shown in Fig.~\ref{fig:framework}, our Hierarchical Auto-Zoom model (HAZN) has three levels of granularity for tackling scale variation in object parsing, \ie image-level, object-level, and part-level. At each level, a fully convolutional neural network (FCN) is used to perform scale/location estimation and part parsing simultaneously. The three levels of FCNs are all built on the same network structure, a modified FCN proposed by~\cite{chen2014semantic}, namely DeepLab-LargeFOV. This network structure is one of the most effective FCNs in segmentation, so we also treat it as our baseline for final performance comparison.

To handle scale variation in objects and parts, the HAZN concatenates two Auto-Zoom Nets (AZNs), namely object-scale AZN and part-scale AZN, into a unified network. The object-scale AZN refines the image-level part score map with object bounding box proposals while the part-scale AZN further refines the object-level part score map with part bounding box proposals. Each AZN employs an auto-zoom process: first estimates the region of interest (ROI), then properly resizes the predicted regions, and finally refine the part scores within the resized regions.

\vspace{-0.7\baselineskip}
\subsection{Object-scale Auto-Zoom Net (AZN)}
\begin{figure}[!t]
\begin{center}
   \includegraphics[width=0.8\linewidth]{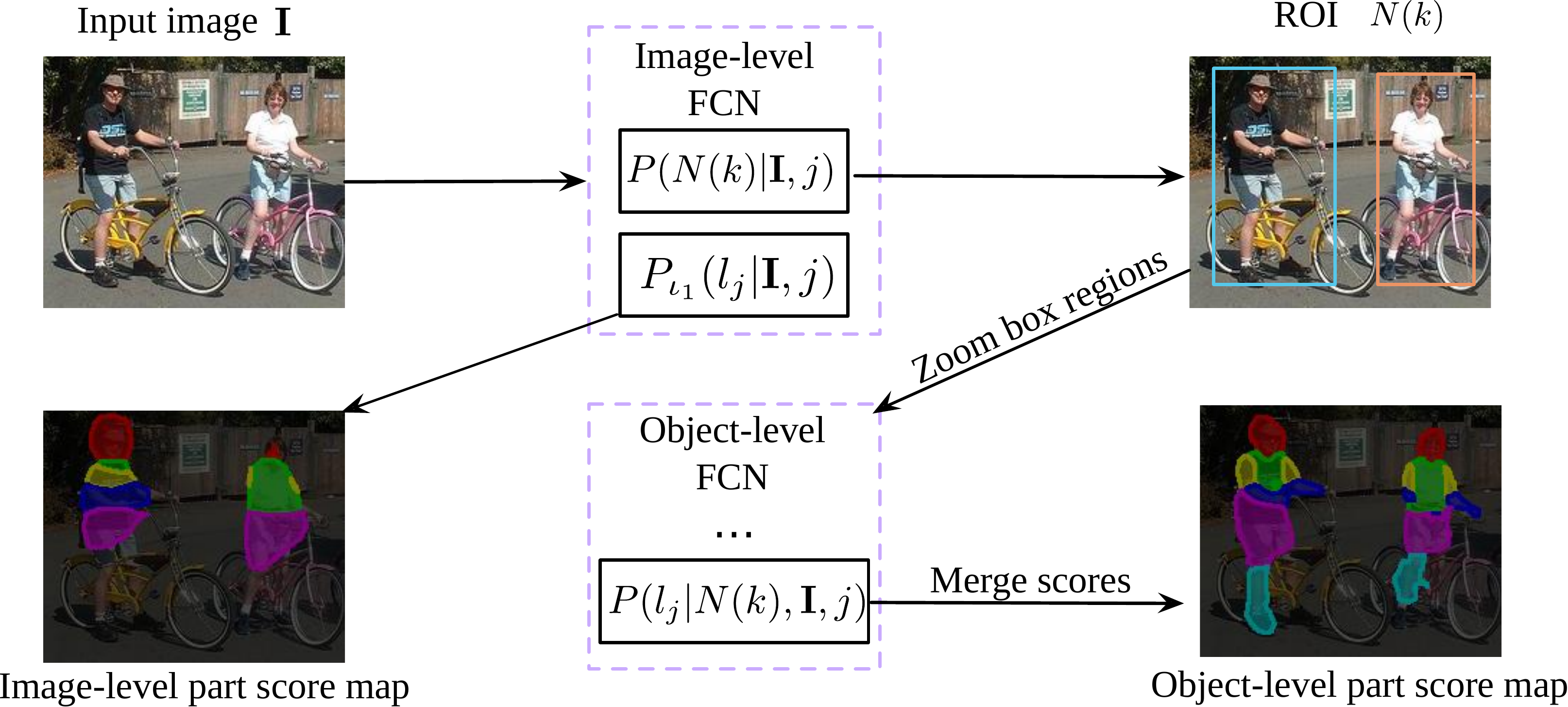}
\end{center}
\vspace{-1\baselineskip}
 \caption{Object-scale Auto-Zoom model from a probabilistic view, which predicts ROI region $N(k)$ at object-scale, and then refines part scores based on the properly zoomed region $N(k)$. Details are in Sec.~\ref{subsec:AZN}.}
\vspace{-1.6\baselineskip}
\label{fig:prob_model}
\end{figure}

\vspace{-0.4\baselineskip}
\label{subsec:AZN}
For the task of object part parsing, we are provided with $n$ training examples $\{\ve{I}_i, \ve{L}_i\}_{i=1}^{n}$, where $\ve{I}$ is the given image and $\ve{L}$ is the supervision information that provides discrete semantic labels of interest. Our target is to learn the posterior distribution $P(l_j | \ve{I},j)$ for each pixel j of an image $\ve{I}$. This distribution is approximated by our object-scale AZN, as shown in Fig.~\ref{fig:prob_model}.

We first use the image-level FCN (see Fig.~\ref{fig:framework}) to produce the image-level part score map $P_{\iota_1}(l_j | \ve{I},j)$, which gives comparable performance to our baseline method (DeepLab-LargeFOV). This is a normal \emph{\textbf{part parsing network}} that uses the original image as input and outputs the pixel-wise part score map. Our object-scale AZN aims to refine this part score map with consideration of object instance scales. To do so, we add a second component to the image-level FCN, performing regression to estimate the size and location of an object bounding box (or ROI) for each pixel, together with a confidence map indicating the likelihood that the box is an object. This component is called a \emph{\textbf{scale estimation network}} (\textbf{SEN}), which shares the first few layers with the part parsing network in the image-level FCN. In math, the SEN corresponds to a probabilistic model $P(b_j | \ve{I}, j)$, where $b_j$ is the estimated bounding box for pixel $j$, and $P(b_j|...)$ is the confidence score of $b_j$.

After getting $\{b_j |\forall j\in\ve{I}\}$, we threshold the confidence map and perform non-maximum suppresion to yield a finite set of object ROIs (typically 5-10 per image, with some overlap): $\{b_k | k \in \ve{I}\}$. Each $b_k$ is associated with a confidence score $P(b_k)$. As shown in Fig.~\ref{fig:framework}, a \textbf{region zooming} operation is then performed on each $b_k$, resizing $b_k$ to a standard-sized ROI $N(k)$. Specifically, this zooming operation computes a zooming ratio $f(b_k, L^{b_k}_p)$ for bounding box $b_k$, based on the bounding box $b_k$ and the computed image-level part labels $L^{b_k}_p$ within the bounding box, and then enlarges or shrinks the image within the bounding box by the zooming ratio. We will discuss $f()$ in detail in Sec.~\ref{sec:exp}.

Now we have a set of zoomed ROI proposals $\{N(k)|k\in\ve{I}\}$, each $N(k)$ associated with score $P(b_k)$. We learn another probabilistic model $P(l_j | N(k), \ve{I}, j)$, which re-estimates the part label for each pixel $j$ within the zoomed ROI $N(k)$. This probabilistic model corresponds to the part parsing network in the object-level FCN (see Fig.~\ref{fig:framework}), which takes as input the zoomed object bounding boxes and outputs the part scores within those object bounding boxes.

The new part scores for the zoomed ROIs need to be merged to produce the object-level part score map for the whole image. Since there may be multiple ROIs that cover a pixel $j$, we define the neighbouring region set for pixel j as $\hua{Q}(j)=\{N(k) | j \in N(k), k \in \ve{I}\}$. Under this definition of $\hua{Q}(j)$, the \textbf{score merging} process can be expressed as Equ.~\ref{eqn:AZN}, which essentially computes the weighted sum of part scores for pixel $j$, from the zoomed ROIs that cover $j$. For a pixel that is not covered by any zoomed ROI, we simply use its image-level part score as the current part score. Formally, the object-level part score $P_{\iota_2}(l_j|\ve{I}, j)$, is computed as, 
\begin{align}
\vspace{-1.6\baselineskip}
P_{\iota_2}(l_j|\ve{I}, j) &= \sum\nolimits_{N(k) \in \hua{Q}(j)} P(l_j | N(k), \ve{I}, j) P(N(k) | \ve{I}, j);  \nonumber \\
P(N(k) | \ve{I}, j) &= P(b_k) / \textstyle{\sum_{k: N(k) \in \hua{Q}(j)} P(b_k)}
\label{eqn:AZN}
\end{align}
\subsection{Hierarchical Auto-Zoom Net (HAZN)}
\vspace{-0.4\baselineskip}
\label{subsec:HAZN}
The scale of object parts can also vary considerably even if the scale of the object is fixed. This leads to a hierarchical strategy with multiple stages, called the Hierarchical Auto-Zoom Net (HAZN), which applies AZNs to images to find objects and then on objects to find parts, followed by a refinement stage. As shown in Fig.~\ref{fig:framework}, we add the part-scale AZN to the end of the object-scale AZN.

We add a second component to the object-level FCN, \ie the SEN network, to estimate the size and location of part bounding boxes, together with confidence maps for every pixel within each zoomed object ROI. Again the confidence map is thresholded, and non-maximal suppresion is applied, to yield a finite set of part ROIs (typically 5-30 per image, with some overlap). Each part ROI is zoomed to a fixed size. Then, we re-estimate the part scores within each zoomed part ROI using the part parsing network in the part-level FCN. The part parsing network is the only component of the part-level FCN, which takes the zoomed part ROI and the zoomed object-level part scores (within the part ROI) as inputs. After getting the part scores within each zoomed Part ROI, the score merging process is the same as in the object-scale AZN.

We can easily extend our HAZN to include more AZNs at finer scale levels if we focus on smaller object parts such as human eyes.
If the HAZN contains $n$ AZNs, there are $n+1$ FCNs needed to be trained. The $\iota$-th FCN learns $P_\iota(l_j | N(j)_{\iota-1}, ...)$ to refine the part scores based on the scale/location estimation results $P_{\iota-1}(N(j)_{\iota-1} | ...)$ and the part parsing results $P_{\iota-1}(l_j | N(j)_{\iota-1}, ...)$ from the previous level $\iota-1$. At the same time, the $\iota$-th FCN also learns $P_\iota(N(j)_l | ...)$ to estimate the region of interest (ROI) for the next level $\iota+1$.

\vspace{0.6\baselineskip}
\subsection{Training and Testing Phases for Object-scale AZN}
\label{subsec:train_test}
In this section, we introduce the specific networks to learn our probalistic models. Specifically, we use a modern version of FCN, \ie \textbf{DeepLab-LargeFOV}~\cite{chen2014semantic}, as our basic network structure. DeepLab-LargeFOV is a stronger variant of DeepLab~\cite{chen2014semantic}, which takes the raw image as input, and outputs dense feature maps. DeepLab-LargeFOV modifies the filter weights at the $fc_6$ layer so that its field-of-view is larger. Due to the space limits, we refer readers to the original paper for details.

\begin{figure}[!t]
\begin{center}
   \includegraphics[width=0.8\linewidth]{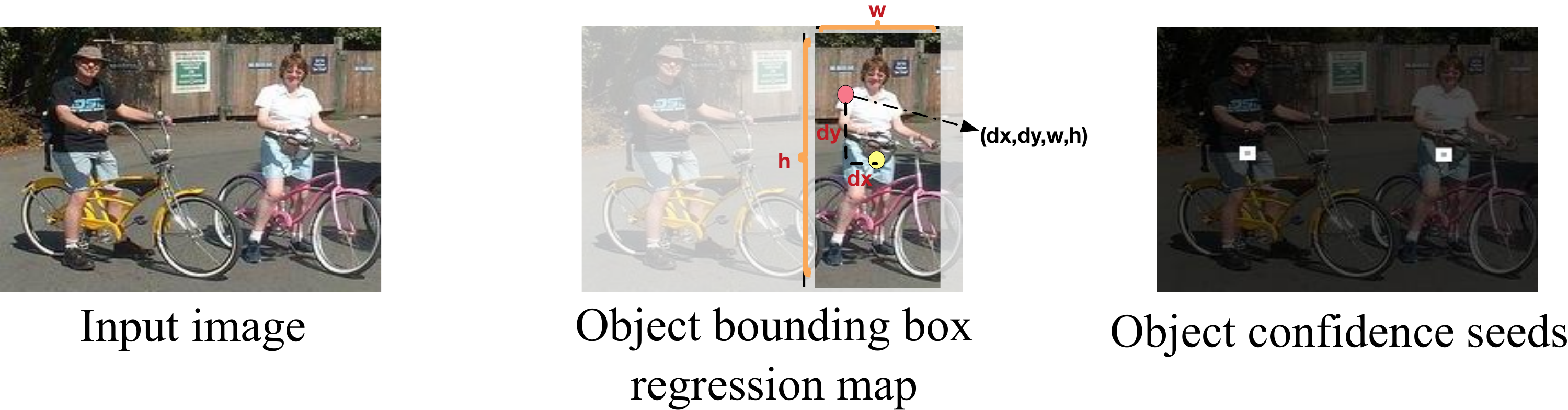}
\end{center}
\vspace{-1\baselineskip}
\caption{Ground truth regression target for training the scale estimation network (SEN) in the image-level FCN. Details in Sec.~\ref{subsec:train_test}.}
\vspace{-1.6\baselineskip}
\label{fig:train_sen}
\end{figure}

\paragraph{Training the SEN.}
The scale estimation network (SEN) aims to regress the region of interest (ROI) for each pixel $j$ in the form of a bounding box, $b_j$. Here we borrow the idea presented in the DenseBox~\cite{huang2015densebox} to do scale estimation, since it is simple and performing well enough for our task. 
In detail, at object level, the ROI of pixel $j$ corresponds to the object instance box that pixel $j$ belongs to. For training the SEN, two output label maps are needed as visualized in Fig.~\ref{fig:train_sen}. The first one is the bounding box regression map $\ve{L}_{b}$, which is a four-channel output for each pixel $j$ to represent its ROI $b_j$: $\ve{l}_{bj}=\{dx_j, dy_j, w_j, h_j\}$. Here $(dx_j, dy_j)$ is the relative position from pixel $j$ to the center of $b_j$; $h_j$ and $w_j$ are the height and width of $b_j$. We then re-scale the outputs by dividing them with $400$. 
The other target output map is a binary confidence seed map $\ve{L}_{c}$, in which $\ve{l}_{cj}\in\{0,1\}$ is the ROI selection indicator at pixel $j$. It indicates the preferred pixels for us to use for ROI prediction, which helps the algorithm prevent many false positives. In practice, we choose the central pixels of each object instance as the confidence seeds, which tend to predict the object bounding boxes more accurately than those pixels at the boundary of an object instance region. 

Given the ground-truth label map of object part parsing, we can easily derive the training examples for the SEN: $\hua{H} = \{\ve{I}_i, \ve{L}_{bi}, \ve{L}_{ci}\}_{i=1}^{n}$, where $n$ is the number of training instances. We minimize the negative log likelihood to learn the weights $\ve{W}$ for the SEN, and the loss $l_{SEN}$ is defined in Equ.~\ref{eqn:loss_sen}.
{\small
\begin{align}
l_{SEN}(\hua{H}|\ve{W}) &= \frac{1}{n}\sum\nolimits_{i}(l_{b}(\ve{I}_i, \ve{L}_{bi} | \ve{W}) + \lambda l_{c}(\ve{I}_i, \ve{L}_{ci} | \ve{W})); \nonumber\\
l_{c}(\ve{I}, \ve{L}_{c}|\ve{W}) &= -\beta\sum\limits_{j:l_{cj}=1} \log P(l_{cj}^*=1|\ve{I}, \ve{W}) -(1-\beta)\sum\limits_{j:l_{cj}=0}\log P(l_{cj}^*=0|\ve{I}, \ve{W}); \nonumber \\
l_{b}(\ve{I}, \ve{L}_{b}|\ve{W}) &= \frac{1}{|\ve{L}_{cj}^+|}\sum\nolimits_{j:l_{cj}=1}\|\ve{l}_{bj}-\ve{l}_{bj}^*\|^2
\label{eqn:loss_sen}
\end{align}
}
For the confidence seeds, we employ the balanced cross entropy loss, where $l_{cj}^*$ and $l_{cj}$ are the predicted value and ground truth value respectively. The probability is from a sigmoid function performing on the activation of the last layer of the CNN at pixel $j$.  $\beta$ is defined as the proportion of pixels with $l_{cj}=0$ in the image, which is used to balance the positive and negative instances. The loss for bounding box regression is the Euclidean distance over the confidence seed points, and $|\ve{L}_{cj}^+|$ is the number of pixels with $l_{cj}=1$.

\paragraph{Testing the SEN.} For testing, the SEN outputs both the confidence score map $P(l_{cj}^*=1|\ve{I}, \ve{W})$ and a four-dimensional bounding box $\ve{l}_{bj}^*$ for each pixel $j$.
We regard a pixel $j$ with confidence score higher than $0.5$ to be reliable and output its bounding box $b_j = \ve{l}_{bj}^*$, associated with confidence score $P(b_j) = P(l_{cj}^*=1|\ve{I}, \ve{W})$. We perform non-maximum suppression based on the confidence scores, yielding several bounding boxes $\{\ve{b}_j | j=1,2,...\}$ as candidate ROIs with confidence scores $P(\ve{b}_j)$. Each candidate ROI $b_j$ is then properly zoomed, becoming $N(j)$.

\paragraph{Training the part parsing.} The training of the part parsing network is standard. For the object-level FCN, the part parsing network is trained based on all the the zoomed image regions (ROIs), with the ground-truth part label maps $\hua{H_p} = \{\ve{L}_{pi}\}_{i=1}^{n}$ within the zoomed ROIs. For the image-level FCN, the part parsing network is trained based on the original training images, and has the same structure as the scale estimation network (SEN) in the FCN. Therefore, we merge the part parsing network with the SEN, yielding the image-level FCN with loss defined in Equ.~\ref{eqn:AZN_loss}. Here,  $l_{p}(\ve{I}, \ve{L}_{p})$ is the commonly used multinomial logistic regression loss for classification.
\begin{align}
l_{AZN}(\hua{H},\hua{H_p}|\ve{W}) =\frac{1}{n}\sum\nolimits_il_{p}(\ve{I}_i, \ve{L}_{pi}) + l_{SEN}(\hua{H}|\ve{W});
\label{eqn:AZN_loss}
\end{align}

\paragraph{Testing the part parsing.}  For testing the object-scale AZN, we first run the image-level FCN, yielding part score maps at the image level and bounding boxes for the object level. Then we zoom onto the bounding boxes and parse these regions based on the object-level FCN model, yielding part score maps at the object level. By merging the part score maps from the two levels, we get better parsing results for the whole image.

\section{Experiments}
\label{sec:exp}
\vspace{-0.4\baselineskip}
\subsection{Implementation Details}
\label{subsec:impDetails}
\paragraph{Selection of confidence seeds.}
To train the scale estimation network (SEN), we need to select confidence seeds for object instances or parts. For human instances, we use the human instance masks from the PASCAL-Person-Part Dataset~\cite{chen2014detect} and select the central $7\times7$ pixels within each instance mask as the confidence seeds. To get the confidence seeds for human parts, we first compute connected part segments from the groundtruth part label map, and then also select the central $7\times7$ pixels within each part segment. We present the details of our approach for humans because the extension to horses and cows is straightforward.
\paragraph{Zooming ratio of ROIs.}
The SEN networks in the FCNs provide a set of human/part bounding boxes (ROIs), $\{b_j | j\in \ve{I}\}$, which are then zoomed to a proper human/part scale. The zooming ratio of $b_j$, $f(b_j, L^{b_j}_p)$, is decided based on the size of $b_j$ and the previously computed part label map $L^{b_j}_p$ within $b_j$. We use slightly different strategies to compute the zooming ratio at the human and part levels. 
For the part level, we simply resize the bounding box to a fixed size, \ie $f_p(b_j) = s_t / max(w_j, h_j)$, where $s_t = 255$ is the target size. Here $w_j$ and $h_j$ are the width and height of $b_j$. 
For the human level, we need to consider the frequently occurred truncation case when only the upper half of a human instance is visible. 
In practice, we use the image-level part label map $L^{b_j}_p$ within the box, and check the existence of legs to decide whether the full body is visible. If the full body is visible, we use the same strategy as parts. Otherwise, we change the target size $s_t$ to $140$, yielding relative smaller region than the full body visible case. We select the target size based on a validation set. 
Finally, we limit all zooming ratio $f_p(b_j)$ within the range $[0.4, 2.5]$ for both human and part bounding boxes to avoid artifacts from up or down sampling of images.

\vspace{-0.7\baselineskip}
\subsection{Experimental Protocol}
\paragraph{Dataset.} We conduct experiments on humans part parsing using the PASCAL-Person-Part dataset annotated by \cite{chen2014detect} which is a subset from the PASCAL VOC 2010 dataset.  The dataset contains detailed part annotations for every person, e.g., head, torso, \etc. We merge the annotations into six clases: Head, Torso, Upper/Lower Arms and Upper/Lower Legs (plus one background class). 
We only use those images containing humans for training (1716 images in the training set) and testing (1817 images in the validation set), the same as~\cite{chen2015attention}. Note that parsing humans in PASCAL is challenging because it has larger variations in scale and pose than other human parsing datasets. 
In addtion, we also perform parsing experiments on the horse-cow dataset~\cite{wang2014semantic}, which contains animal instances in a rough bounding box. In this dataset, we keep the same experimental setting with~\cite{wang2015joint}.

\paragraph{Training.} We train the FCNs using stochastic gradient descent with mini-batches. Each mini-batch contains
30 images. The initial learning rate is 0.001 (0.01 for the final classifier layer) and is decreased by a factor of 0.1 after every 2000 iterations. We set the momentum to be 0.9 and the weight decay to be 0.0005. The initialization model is a modified VGG-16 network pre-trained on ImageNet. Fine-tuning our network on all the reported experiments takes about 30 hours on a NVIDIA Tesla K40 GPU. After training, the average inference time for one PASCAL image is 1.3 s/image.

\paragraph{Evaluation metric.} The object parsing results is evaluated in terms of mean IOU (mIOU). It is computed as the pixel intersection-over-union (IOU) averaged across classes \cite{everingham2014pascal}, which is also adopted recently to evaluate parts~\cite{wang2015joint,chen2015attention}. We also evaluate the part parsing performance \wrt each object instance in terms of $AP^r_{part}$ as defined in~\cite{hariharan2014hypercolumns}.

\paragraph{Network architecture.} We use DeepLab-LargeFOV~\cite{chen2014semantic} as building blocks for the FCNs in our Hierarchical Auto-Zoom Net (HAZN). Recall that our HAZN consists of three FCNs working at different levels of granularity: image level, object level, and part level. At each level, HAZN outputs part parsing scores, and estimats locations and scales for the next level of granularity (\eg objects or parts).

\vspace{-0.7\baselineskip}
\subsection{Experimental Results on Parsing Humans in the Wild}
\label{subsec:human_parsing}
\begin{table}[!t]
  \centering
\setlength{\tabcolsep}{5pt}
\resizebox{1\columnwidth}{!}{
\begin{tabular}{l | c c c c c c c | c}
\toprule[0.2 em]
Method & head & torso & u-arms & l-arms & u-legs & l-legs & bg & Avg \\ \midrule \midrule
DeepLab-LargeFOV~\cite{chen2014semantic} & 78.09 & 54.02 & 37.29 & 36.85 & 33.73 & 29.61 & 92.85 & 51.78 \\
DeepLab-LargeFOV-CRF & 80.13 & 55.56 & 36.43  & 38.72 & 35.50 & 30.82  & 93.52 & 52.95 \\
Multi-Scale Averaging & 79.89 & 57.40 & 40.57 & 41.14 & 37.66 & 34.31 & 93.43 & 54.91 \\
Multi-Scale Attention~\cite{chen2015attention} & \textbf{81.47} & 59.06 & 44.15 & 42.50 & 38.28 & 35.62 & 93.65 & 56.39 \\
\hline 
HAZN (no object scale) & 80.25 & 57.20 & 42.24 & 42.02 & 36.40 & 31.96 & 93.42 & 54.78 \\
HAZN (no part scale) & 79.83 & 59.72 & 43.84 & 40.84 & 40.49 & 37.23 & 93.55 & 56.50 \\
HAZN (full model) & 80.76 & \textbf{60.50} & \textbf{45.65} & \textbf{43.11} & \textbf{41.21} & \textbf{37.74} & \textbf{93.78} & \textbf{57.54} \\
\bottomrule[0.1 em]
\end{tabular}
}
\caption{Part parsing accuracy (\%) on PASCAL-Person-Part in terms of mean IOU. We compare our full model (HAZN) with two sub-models and four state-of-the-art baselines.}
\label{table:seg_eval}
\vspace{-1.7\baselineskip}
\end{table}

\paragraph{Comparison with state-of-the-arts.}
As shown in \tabref{table:seg_eval}, we compare our full model (HAZN) with four baselines. The first one is DeepLab-LargeFOV~\cite{chen2014semantic}. The second one is DeepLab-LargeFOV-CRF, which adds a post-processing step to DeepLab-LargeFOV by means of a fully-connected Conditional Random Field (CRF)~\cite{KrahenbuhlK11}. CRFs are commonly used as postprocessing for object semantic segmentation to refine boundaries~\cite{chen2014semantic}. The third one is Multi-Scale Averaging, which feeds the DeepLab-LargeFOV model with images resized to three fixed scales (0.5, 1.0 and 1.5) and then takes the average of the three part score maps to produce the final parsing result. The fourth one is Multi-Scale Attention~\cite{chen2015attention}, a most recent work which uses a scale attention model to handle the scale variations in object parsing.

Our HAZN obtains the performance of 57.5\%, which is 5.8\% better than DeepLab-LargeFOV, and 4.5\% better than DeepLab-LargeFOV-CRF. Our model significantly improves the segmentation accuracy in all parts. Note we do not use any CRF for post processing. The CRF, though proven effective in refining boundaries in object segmentation, is not strong enough at recovering details of human parts as well as correcting the errors made by the DeepLab-LargeFOV. 

The third baseline (Multi-Scale Averaging) enumerates multi-scale features which is commonly used to handle the scale variations, yet its performance is poorer than ours, indicating the effectiveness of our Auto-Zoom framework.

Our overall mIOU is 1.15\% better than the fourth baseline (Multi-Scale Attention), but we are much better in terms of detailed parts like upper legs (around 3\% improvement). In addition, we further analyze the scale-invariant ability in Tab.~\ref{table:seg_eval_scalewise}, which both methods aim to improve. We can see that our model surpasses Multi-Scale Attention in all instance sizes especially at size XS (9.5\%) and size S (5.5\%). 

\paragraph{Importance of object and part scale.}
As shown in \tabref{table:seg_eval}, we study the effect of the two scales in our HAZN.
In practice, we remove either  the object-scale AZN or the part-scale AZN from the full HAZN model, yielding two sub-models: (1) \textbf{HAZN (no object scale)}, which only handles the scale variation at part level. (2)\textbf{HAZN (no part scale)}, which only handles the scale variation at object instance level. 

Compared with our full model, removing the object-scale AZN causes 2.8\% mIOU degradation while removing the part-scale AZN results in 1\% mIOU degradation. 
We can see that the object-scale AZN, which handles the scale variation at object instance level, contributes a lot to our final parsing performance. For the part-scale AZN, it further improves the parsing by refining the detailed part predictions, \eg around 3\% improvement of lower arms as shown in \tabref{table:seg_eval}, yielding visually more satisfactory results. This demonstrates the effectiveness of the two scales in our HAZN model.

\begin{table}[!b]
  \centering
  \setlength{\tabcolsep}{10pt}
   \resizebox{0.9\columnwidth}{!}{
  \begin{tabular}{l | c c c c}
    \toprule[0.2 em]
    Method & Size XS & Size S & Size M & Size L \\ \midrule \midrule
    DeepLab-LargeFOV~\cite{chen2014semantic} & 32.5 & 44.5 & 50.7 & 50.9 \\
    DeepLab-LargeFOV-CRF & 31.5 & 44.6 & 51.5 & 52.5 \\
    Multi-Scale Averaging & 33.7 & 45.9 & 52.5 & 54.7 \\
    Multi-Scale Attention~\cite{chen2015attention} & 37.6 & 49.8 & 55.1 & 55.5 \\
    \hline 
    HAZN (no object scale) & 38.2 & 51.0 & 55.1 & 53.4 \\    
    HAZN (no part scale) & 45.1 & 53.1 & 55.0 & 55.0  \\
    HAZN (full model) & \textbf{47.1} & \textbf{55.3} & \textbf{56.8} & \textbf{56.0} \\ \bottomrule[0.1 em]
  \end{tabular}
  }
 \caption{Part parsing accuracy w.r.t. size of human instance (\%) on PASCAL-Person-Part in terms of mean IOU.}
\vspace{-1.\baselineskip}
\label{table:seg_eval_scalewise}
\end{table}  

\paragraph{Part parsing accuracy w.r.t. size of human instance.} Since we handle human with various sizes, it is important to check how our model performs with respect to the change of human size in images. 
In our experiments, we categorize all the ground truth human instances into four different sizes according to the bounding box area of each instance $s_b$ (the square root of the bounding box area). Then we compute the mean IOU (within the bounding box) for each of these four scales.

The four sizes are defined as follows: (1) Size XS: $s_b \in [0,80]$, where the human instance is extremely small in the image; (2) Size S: $s_b \in [80,140]$; (3) Size M: $s_b \in [140,220]$; (4) Size L: $s_b \in [220,520]$, which usually corresponds to truncated human instances where the human's head or torso covers the majority of the image. 

The results are given in \tabref{table:seg_eval_scalewise}. The baseline DeepLab-LargeFOV performs badly at size XS or S (usually only the head or the torso can be detected by the baseline), while our HAZN (full model) improves over it significantly by 14.6\% for size XS and 10.8\% for size S. This shows that HAZN is particularly good for small objects, where the parsing is difficult to obtain. For instances in size M and L, our model also significantly improve the baselines by around 5\%. In general, by using HAZN, we achieve much better scale invariant property to object size than a generally used FCN type of model. We also list the results for the other three baselines for reference. 

In addition, it is also important to jointly perform the two scale AZNs in a sequence. To show this, we additionally list the results from our model without object/part scale AZN in the $5_{th}$ and the $6_{th}$ row respectively. 
By jumping over object scale (HAZN no object scale), the performance becomes significantly worse at size XS, since the model can barely detect the object parts at the image-level when the object is too small.  However, if we remove part scale (HAZN no part scale), the performance also dropped in all sizes. This is because using part scale AZN can recover the part details much better than only using object scale. 
Our HAZN (full model), which sequentially leverage the benefits from both the object scale and part scale, yielding the best performance overall.

\paragraph{Instance-wise part parsing accuracy.} We evaluate our part parsing results \wrt each human instance in terms of $AP^r_{part}$ as defined in~\cite{hariharan2014hypercolumns}. The segment IOU threshold is set to 0.5. A human instance segment is correct only when it overlaps enough with a groundtruth instance segment. To compute the intersection of two segments, we only consider the pixels whose part labels are also right. 

To generate instance segmentation (which is not our major task), we follow a similar strategy to~\cite{hariharan2014hypercolumns} by first generating object detection box and then doing instance segmentation. Specifically, we use fast R-CNN to produce a set of object bounding box proposals, and each box is associated with a confidence score. Then within each bounding box, we use FCN to predict a coarse object instance mask, and use the coarse instance mask to retrieve corresponding part segments from our final HAZN part label map. Last, we use the retrieved part segments to compose a new instance mask where we keep the boundary of part segments. In the instance overlapping cases, we follow the boundary from the predicted instance mask.

We first directly compare with the number reported by~\cite{hariharan2014hypercolumns}, on the whole validation set of PASCAL 2010. Our full HAZN achieves \textbf{43.08\%} in $AP^r_{part}$, \textbf{14\%} higher than~\cite{hariharan2014hypercolumns}. We also compare with two state-of-the-art baselines (DeepLab-LargeFOV~\cite{chen2014semantic} and Multi-Scale Attention~\cite{chen2015attention}) on the PASCAL-Person-Part dataset. For both baselines, we applied the same strategy to generate instances but with different part parsing results. As shown in Tab.\ref{table:apr_part}, our model is 12\% points higher than DeepLab-LargeFOV and 6\% points higher than Multi-Scale Attention, in terms of $AP^r_{part}$. 

\begin{table}[!b]
  \centering
  \setlength{\tabcolsep}{10pt}
   \resizebox{0.8\columnwidth}{!}{
  \begin{tabular}{l | c c c}
    \toprule[0.2 em]
        & DeepLab-LargeFOV~\cite{chen2014semantic} & Multi-Scale Attention~\cite{chen2015attention} &  HAZN(full model) \\ \midrule \midrule
    $AP^r_{part}$ & 31.32 & 37.53 & 43.72 \\
 \bottomrule[0.1 em]
  \end{tabular}
  }
  \caption{Instance-wise part parsing accuracy on PASCAL-Person-Part in terms of $AP^r_{part}$.}
 \vspace{-1.\baselineskip}
  \label{table:apr_part}
\end{table}  

\begin{figure} [t]
\vspace{-0.1\baselineskip}
\centering
\includegraphics[width=1.00\linewidth]{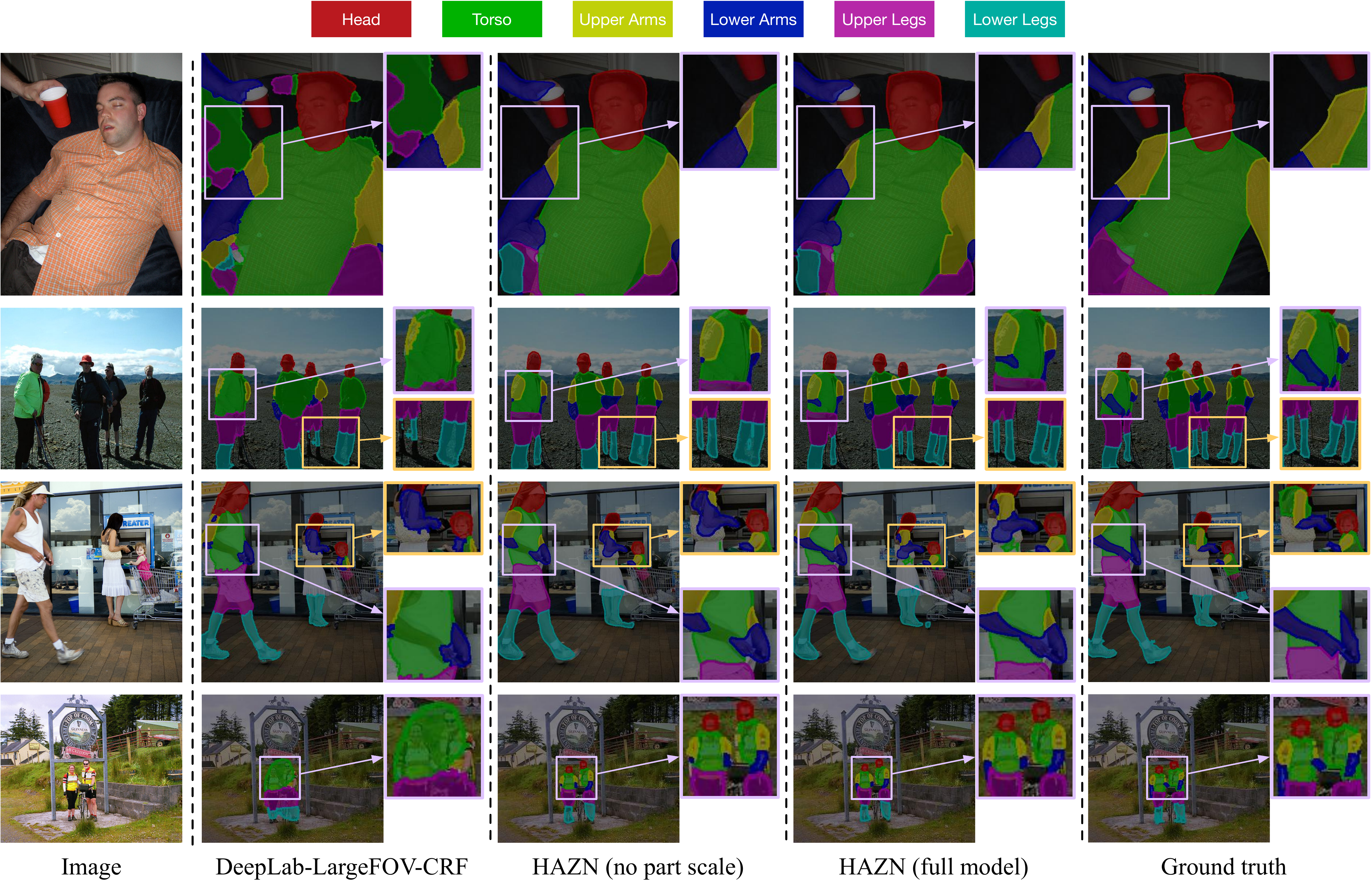}
\vspace{-1.0\baselineskip}
\caption{Qualitative comparison on the PASCAL-Person-Part dataset. We compare with DeepLab-LargeFOV-CRF~\cite{chen2014semantic} and HAZN (no part scale). Our proposed HAZN models (the $3_{rd}$ and $4_{th}$ columns) attain better visual parsing results, especially for small scale human instances and small parts such as legs and arms.}
\vspace{-1.6\baselineskip}
\label{fig:seg_image}
\end{figure}

\paragraph{Qualitative results} We visually show several example results from the PASCAL-Person-Part dataset in \figref{fig:seg_image}. The baseline DeepLab-LargeFOV-CRF produces several errors due to lack of object and part scale information, \eg background confusion ($1_{st}$ row), human part confusion ($3_{rd}$ row), or important part missing ($4_{th}$ row), \etc, yielding non-satisfactory part parsing results. Our HAZN (no part scale), which only contains object-scale AZN, already successfully relieves the confusions for large scale human instances while recovers the parts for small scale human instances. By further introducing part scale, the part details and boundaries are recovered even better, which are more visually satisfactory.

\begin{figure}[!htbp]
\centering
\hspace*{-0.4cm}
\includegraphics[width=1.05\linewidth]{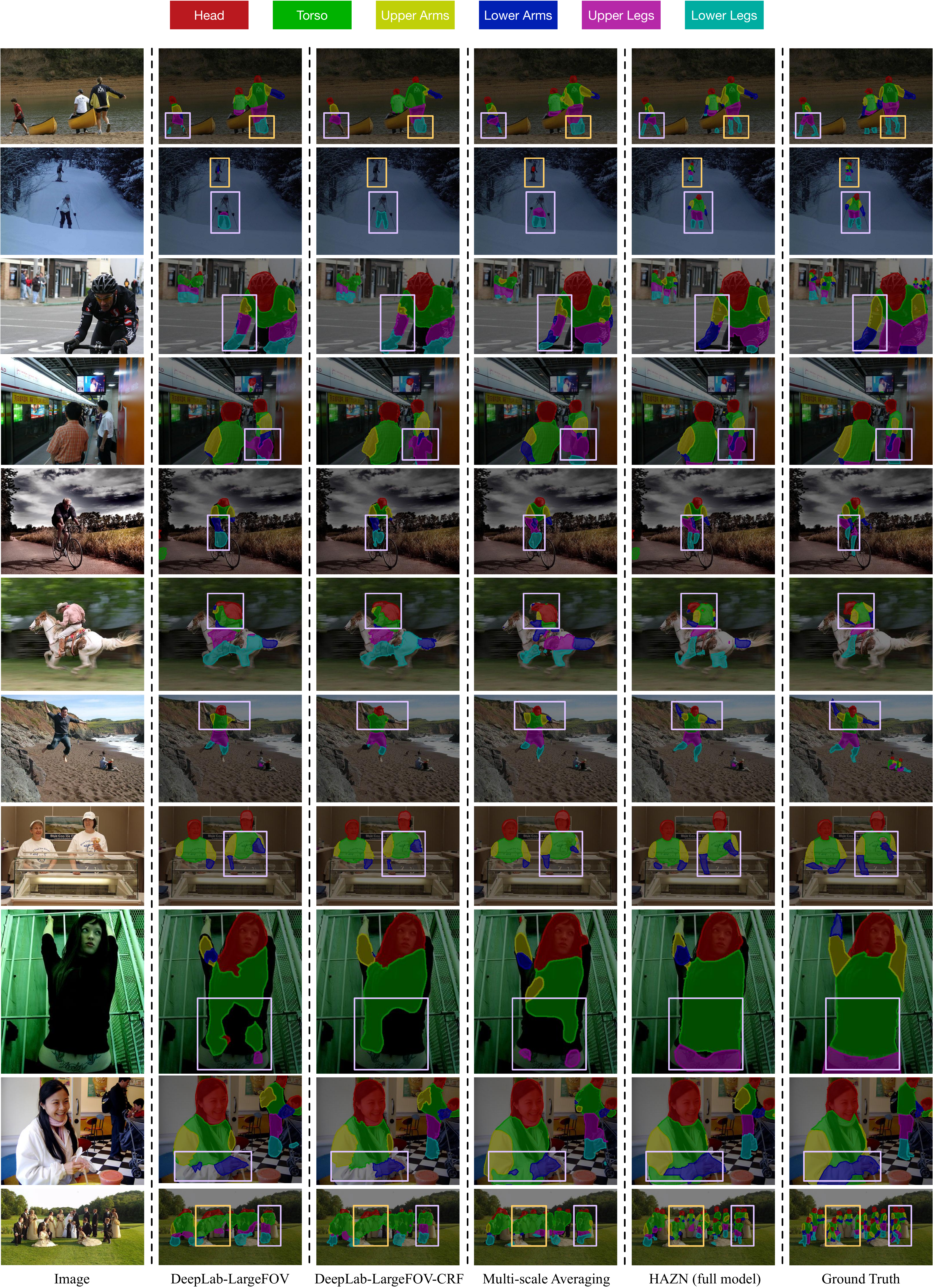}
\caption{More qualitative comparison on PASCAL-Person-Part. The baselines are explained in Sec.~\ref{subsec:human_parsing}.}
\label{fig:human_res}
\end{figure}

More visual examples are provided in Fig.~\ref{fig:human_res}, comparing with more baselines. It can be seen that our full model (HAZN) gives much more satisfied part parsing results than the state-of-the-art baselines. Specifically, for small-scale human instances (\eg the 1, 2, 5 rows of the figure), our HAZN recovers human parts like lower arms and lower legs and gives more accurate part boundaries; for medium-scale or large-scale human instances (\eg the 3, 4, 9, 10 rows of the figure), our model relieves the local confusion with other parts or with the background.

\begin{figure}[!t]
 \centering
\includegraphics[width=1.0\linewidth]{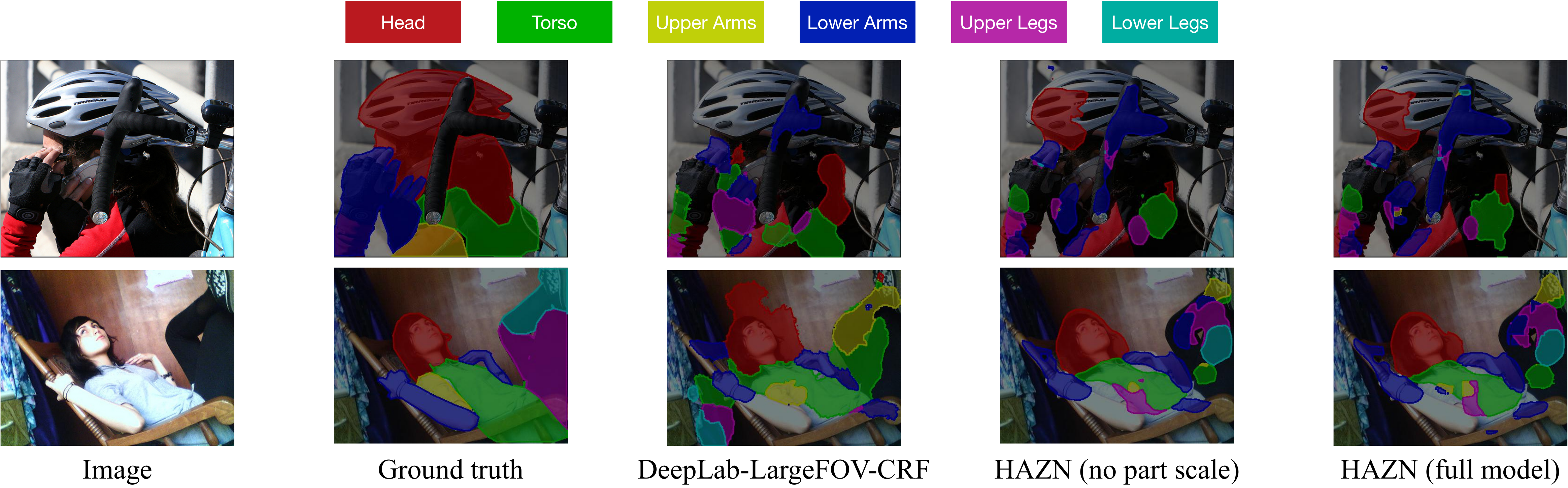}
\caption{Failure cases for both the baseline and our models.}
\vspace{-1.2\baselineskip}
\label{fig:seg_image_fail}
\end{figure}

\paragraph{Failure cases.} Our typical failure modes are shown in \figref{fig:seg_image_fail}. Compared with the baseline DeepLab-LargeFOV-CRF, our models give more reasonable parsing results with less local confusion, but they still suffer from heavy occlusion and unusual poses.

\subsection{Experiments on the Horse-Cow Dataset}
\label{sec:horses}
\vspace{-0.3\baselineskip}
To show the generality of our method to instance-wise object part parsing, we also applied our method to horse instances and cow instances presented in~\cite{wang2014semantic}. All the testing procedures are the same as those described above for humans.
 
We copy the baseline numbers from~\cite{wang2015joint}, and give the evaluation results in Tab.~\ref{table:part_bbox_eval}. It shows that our baseline models from the DeepLab-LargeFOV~\cite{chen2014semantic} already achieve competative results with the state-of-the-arts, while our HAZN provides a big improvement for horses and cows. The improvement over the state-of-the-art method \cite{wang2015joint} is roughly 5\% mIOU. It is most noticeable for small parts, e.g. the improvement for detecting horse/cow head and cow tails is more than 10$\%$. This shows that our auto-zoom strategy can be effectively generalized to other objects for part parsing. 

\begin{figure}[!htbp]
\centering
\includegraphics[width=1.00\linewidth]{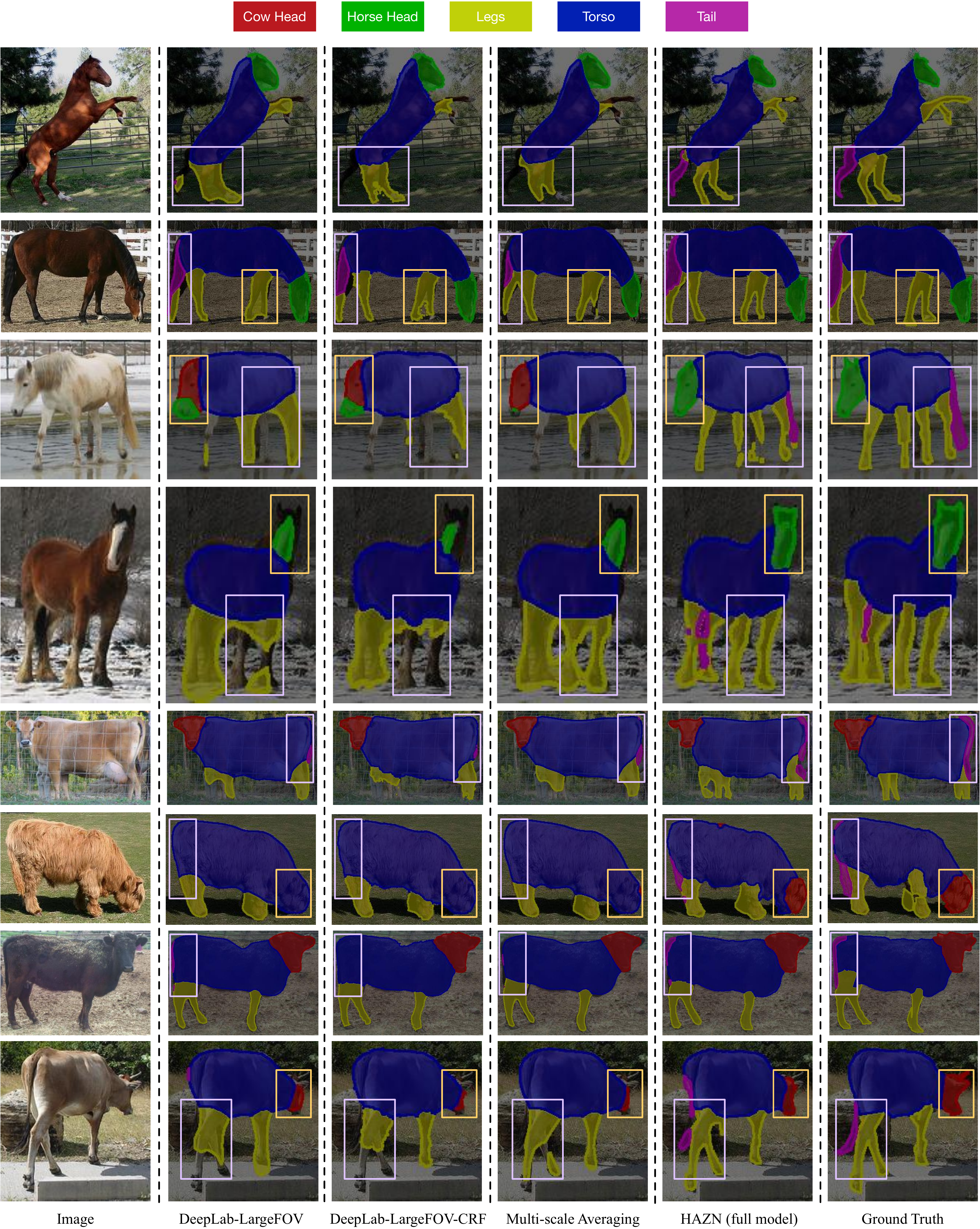}
\vspace{-1.0\baselineskip}
\caption{Qualitative comparison on the Horse-Cow Dataset. The baselines are explained in Sec.~\ref{subsec:human_parsing}.}
\label{fig:horse_cow_res}
\end{figure}

We also provide qualitative evaluations in Fig.~\ref{fig:horse_cow_res}, comparing our full model with three state-of-the-art baselines. The three baselines are explained in Sec.~\ref{subsec:human_parsing}. We can observe that using our model, small parts such as legs and tails have been effectively recovered, and the boundary accuracy of all parts has been improved. 

\begin{table}[!htpb] %
\vspace{-1\baselineskip}
  \setlength{\tabcolsep}{5pt}
  \resizebox{1.0\columnwidth}{!}{
  \begin{tabular}{l |c c c c c | c || c c c c c | c}
 \toprule[0.2 em]
	\multicolumn{7}{c}{Horse} &  \multicolumn{6}{c}{Cow}\\
     \midrule
    Method & Bkg & head & body & leg & tail &  Avg. &  Bkg & head & body & leg & tail &  Avg. \\ \midrule 
	SPS~\cite{wang2014semantic} & 79.14 & 47.64 & 69.74 & 38.85 & -  & - & 78.00 & 40.55 & 61.65 & 36.32 & -  & - \\
	HC$^*$~\cite{hariharan2014hypercolumns} & 85.71 & 57.30 & {77.88} & 51.93 & 37.10  & 61.98 & 81.86 & 55.18 & 72.75 & 42.03 & 11.04  & 52.57 \\
	JPO~\cite{wang2015joint} & {87.34} & {60.02} & 77.52 & {58.35} & \textbf{51.88}  & {67.02} & {85.68} & {58.04} & {76.04} & {51.12} & {15.00} & {57.18} \\
	\midrule
         LargeFOV & 87.44 & 64.45 & 80.70 & 54.61 & 44.03 & 66.25  & 86.56 & 62.76 & 78.42 & 48.83 & 19.97 & 59.31 \\
	HAZN & \textbf{90.94}  & \textbf{70.75}    &  \textbf{84.49}  & \textbf{63.91}  & 51.73 & \textbf{72.36}    & \textbf{90.71}  & \textbf{75.18}  &  \textbf{83.33}  & \textbf{57.42}  & \textbf{29.37}  &  \textbf{67.20}\\
\bottomrule[0.1 em]
  \end{tabular}
}
  \caption{Mean IOU (mIOU) over the Horse-Cow dataset. We compare with the semantic part segmentation (SPS)~\cite{wang2014semantic}, the Hypercolumn (HC$^*$)~\cite{hariharan2014hypercolumns} and the joint part and object (JPO) results~\cite{wang2015joint}. We also list the performance of DeepLab-LargeFOV (LargeFOV)~\cite{chen2014semantic}. }
\label{table:part_bbox_eval}
\vspace{-1\baselineskip}
\end{table}

\vspace{-1.5\baselineskip}
\section{Conclusions}
\vspace{-0.7\baselineskip}
In this paper, we propose the ``Hierarachical Auto-Zoom Net'' (HAZN) to parse objects in the wild, yielding per-pixel segmentation of the object parts. It adaptably estimates the scales of objects, and their parts, by a two-stage process of Auto-Zoom Nets. We show that on the challenging PASCAL dataset, HAZN performs significantly better (by 5\% mIOU), compared to state of the art methods, when applied to humans, horses, and cows. Unlike standard methods which process the image at a fixed range of scales, HAZN's strategy of searching for objects and then for parts enables it, for example, to zoom in to small image regions and enlarge them to scales which would be prohibitively expensive (in terms of memory) if applied to the entire image (as fixed scale methods would require).

In the future, we would love to extend our HAZN to parse more detailed parts, such as human hand and human eyes. Also, the idea of our AZN can be applied to other tasks like pose estimation in the wild, to make further progress.


{
\bibliography{eccv2016}
\bibliographystyle{splncs03}
}

\end{document}